\documentclass[11pt,a4paper]{article}
\usepackage[hyperref]{acl2018}
\usepackage{times}
\usepackage{latexsym}
\usepackage{graphicx}
\usepackage{tabularx}
\usepackage{multirow}
\usepackage{soul}
\usepackage{color}
\usepackage{graphicx}
\usepackage{tabularx}
\usepackage{multirow}
\usepackage{soul}
\usepackage{color}
\usepackage{url}
\usepackage[ruled]{algorithm2e}

\aclfinalcopy 
\def\aclpaperid{14} 

\title{\textbf{\textsc{Yedda}: A Lightweight Collaborative Text Span Annotation Tool}}

\author{Jie Yang, Yue Zhang, Linwei Li, Xingxuan Li \\
  Singapore University of Technology and Design\\
  {\tt \{jie\_yang, linwei\_li, xingxuan\_li\}@mymail.sutd.edu.sg} \\
  {\tt yue\_zhang@sutd.edu.sg} \\
  }

\date{}

\begin{document}
\maketitle
\begin{abstract}
  In this paper, we introduce \textsc{Yedda}, a lightweight but efficient and comprehensive open-source tool for text span annotation. \textsc{Yedda} provides a systematic solution for text span annotation, ranging from collaborative user annotation to administrator evaluation and analysis. It overcomes the low efficiency of traditional text annotation tools by annotating entities through both command line and shortcut keys, which are configurable with custom labels. \textsc{Yedda} also gives intelligent recommendations by learning the up-to-date annotated text. An administrator client is developed to evaluate annotation quality of multiple annotators and generate detailed comparison report for each annotator pair. Experiments show that the proposed system can reduce the annotation time by half compared with existing annotation tools. And the annotation time can be further compressed by 16.47\% through intelligent recommendation. 
\end{abstract}

\section{Introduction}

Natural Language Processing (NLP) systems rely on large-scale training data \cite{marcus1993building} for supervised training. However, manual annotation can be time-consuming and expensive. Despite detailed annotation standards and rules, inter-annotator disagreement is inevitable because of human mistakes, language phenomena which are not covered by the annotation rules and the ambiguity of language itself \cite{plank2014linguistically}. 

Existing annotation tools \cite{cunningham2002gate,morton2003wordfreak,chen2013anafora,druskat2014atomic} mainly focus on providing a visual interface for user annotation process but rarely consider the post-annotation quality analysis, which is necessary due to the inter-annotator disagreement. In addition to the annotation quality, efficiency is also critical in large-scale annotation task, while being relatively less addressed in existing annotation tools \cite{ogren2006knowtator,stenetorp2012brat}. Besides, many tools \cite{ogren2006knowtator,chen2013anafora} require a complex system configuration on either local device or server, which is not friendly to new users.

\begin{figure}[t] 
  \centering 
  \includegraphics[width=3in]{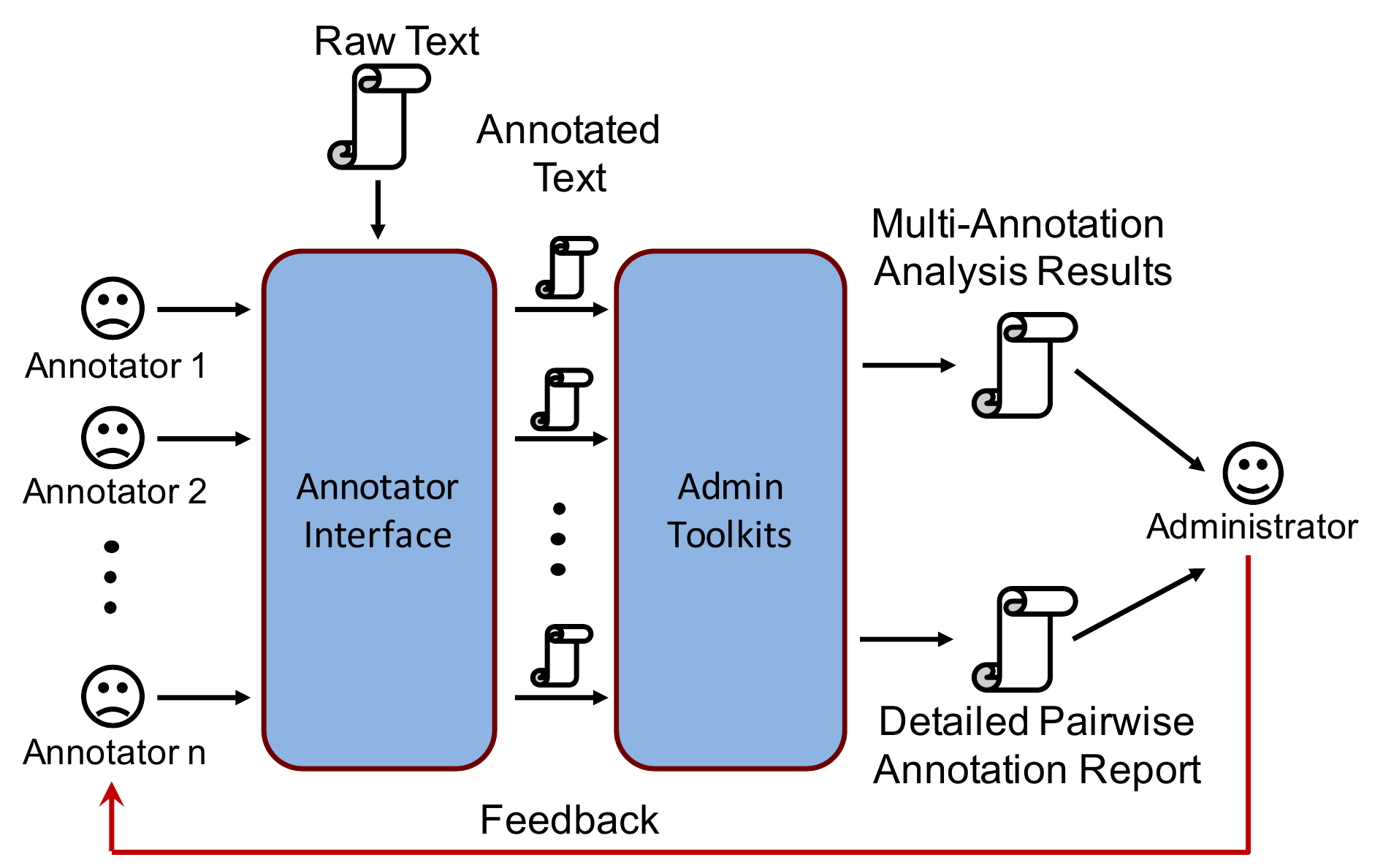}
  \caption{Framework of \textsc{Yedda}.}
  \label{fig:framework}
\end{figure}

\begin{table*}[!t]
\centering
\resizebox{\textwidth}{!}{%
\begin{tabular}{|c|ccc|c|c|c|c|c|c|}
\hline
 \multirow{2}*{Tool}&  \multicolumn{3}{|c|}{Operating System}&\multirow{2}*{\shortstack[c]{Self \\Consistency}}&\multirow{2}*{\shortstack[c]{Command Line\\ Annotation}}&\multirow{2}*{\shortstack[c]{System\\ Recommendation}}&\multirow{2}*{Analysis} &\multirow{2}*{Size}&\multirow{2}*{Language} \\
\cline{2-4}
 &   MacOS&Linux& Win& &&&&&\\
\hline
WordFreak&  ${\surd}$& ${\surd}$& ${\surd}$& ${\surd}$ &${\times}$& ${\surd}$& ${\times}$ &1.1M & Java\\
\hline
\textsc{Gate} &  ${\surd}$& ${\surd}$& ${\surd}$&${\surd}$& ${\times}$ & ${\surd}$& ${\times}$ & 544M &Java\\
\hline
Knowtator&  ${\surd}$& ${\surd}$& ${\surd}$ &${\times}$ & ${\times}$& ${\times}$& ${\surd}$ & 1.5M  &Java \\
\hline
Stanford&  ${\surd}$& ${\surd}$&${\surd}$ &${\surd}$ & ${\times}$& ${\times}$& ${\times}$ & 88k  &Java \\
\hline
Atomic&${\surd}$  &${\surd}$ &${\surd}$ &${\surd}$ & ${\times}$  &${\times}$  & ${\times}$ & 5.8M  &Java \\
\hline
WebAnno&${\surd}$  &${\surd}$ &${\surd}$  &${\times}$ &${\times}$  &${\surd}$  &${\surd}$  &13.2M  &Java  \\
\hline
Anafora&${\surd}$  &${\surd}$ &${\times}$  &${\times}$ & ${\times}$ &${\times}$  & ${\times}$ &1.7M  & Python \\
\hline
\textsc{Brat} & ${\surd}$&${\surd}$& ${\times}$&${\times}$& ${\times}$ & ${\surd}$& ${\times}$& 31.1M  & Python \\
\hline
\hline
\textsc{Yedda} & ${\surd}$&${\surd}$& ${\surd}$ &${\surd}$&${\surd}$ & ${\surd}$& ${\surd}$& 80k & Python \\
\hline
\end{tabular}
}
\caption{Annotation Tool Comparison .} \label{tab:related}
\end{table*}

\begin{figure*}[t] 
  \centering 
  \includegraphics[width=\textwidth]{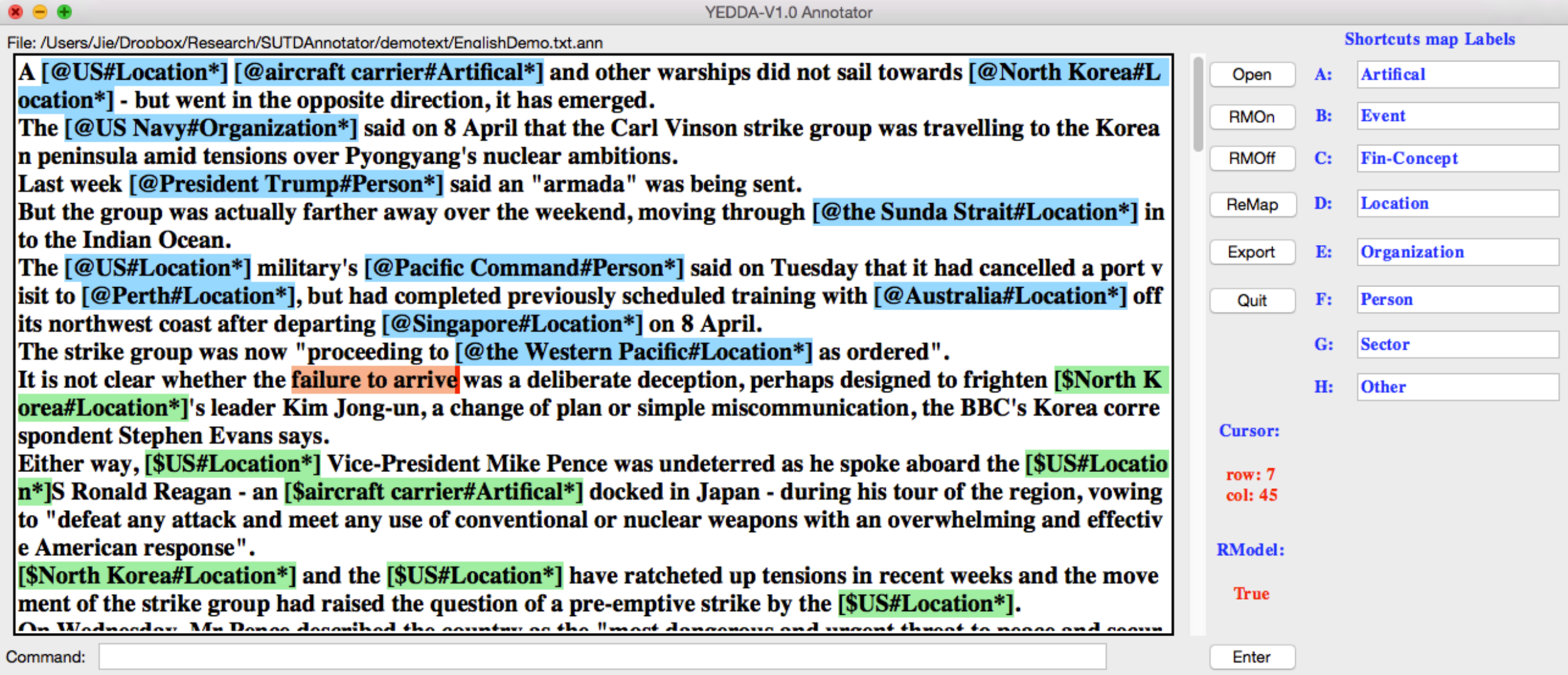}
  \caption{Annotator Interface.}
  \label{fig:annotatorInterface}
\end{figure*}

To address the challenges above, we propose \textsc{Yedda}\footnote{Code is available at \url{https://github.com/jiesutd/YEDDA}.}
, a lightweight and efficient annotation tool for text span annotation. A snapshot is shown in Figure \ref{fig:annotatorInterface}. Here text span boundaries are selected and assigned with a label, which can be useful for Named Entity Recognition (NER) \cite{tjong2003introduction}, word segmentation \cite{sproat2003first}, chunking \cite{tjong2000introduction} ,etc. To keep annotation efficient and accurate, \textsc{Yedda} provides systematic solutions across the whole annotation process, which includes the shortcut annotation, batch annotation with a command line, intelligent recommendation, format exporting and administrator evaluation/analysis.

Figure \ref{fig:framework} shows the general framework of \textsc{Yedda}. It offers annotators with a simple and efficient Graphical User Interface (GUI) to annotate raw text. For the administrator, it provides two useful toolkits to evaluate multi-annotated text and generate detailed comparison report for annotator pair. \textsc{Yedda} has the advantages of being:

\noindent \textbullet $\,$ \textbf{Convenient}: it is lightweight with an intuitive interface and does not rely on specific operating systems or pre-installed packages. 
 
\noindent \textbullet $\,$ \textbf{Efficient}: it supports both shortcut and command line annotation models to accelerate the annotating process.

\noindent \textbullet $\,$ \textbf{Intelligent}: it offers user with real-time system suggestions to avoid duplicated annotation.

\noindent \textbullet $\,$ \textbf{Comprehensive}: it integrates useful toolkits to give the statistical index of analyzing multi-user annotation results and generate detailed content comparison for annotation pairs.

This paper is organized as follows: Section 2 gives an overview of previous text annotation tools and the comparison with ours. Section 3 describes the architecture of \textsc{Yedda} and its detail functions. Section 4 shows the efficiency comparison results of different annotation tools. Finally, Section 5 concludes this paper and give the future plans.

\section{Related Work}
There exists a range of text span annotation tools which focus on different aspects of the annotation process. Stanford manual annotation tool\footnote{\url{http://nlp.stanford.edu/software/stanford-manual-annotation-tool-2004-05-16.tar.gz}} is a lightweight tool but does not support result analysis and system recommendation. Knowtator \cite{ogren2006knowtator} is a general-task annotation tool which links to a biomedical onto ontology to help identify named entities and relations. It supports quality control during the annotation process by integrating simple inter-annotator evaluation, while it cannot figure out the detailed disagreed labels. WordFreak \cite{morton2003wordfreak} adds a system recommendation function and integrates active learning to rank the unannotated sentences based on the recommend confidence, while the post-annotation analysis is not supported.

Web-based annotation tools have been developed to build operating system independent annotation environments.  \textsc{Gate}\footnote{\textsc{Gate} is a general NLP tool which includes annotation function.} \cite{bontcheva2013gate} includes a web-based with collaborative annotation framework which allows users to work collaboratively by annotating online with shared text storage. \textsc{Brat} \cite{stenetorp2012brat} is another web-based tool, which has been widely used in recent years, it provides powerful annotation functions and rich visualization ability, while it does not integrate the result analysis function. Anafora \cite{chen2013anafora} and Atomic \cite{druskat2014atomic} are also web-based and lightweight annotation tools, while they don't support the automatic annotation and quality analysis either. WebAnno \cite{yimam2013webanno,de2016web} supports both the automatic annotation suggestion and annotation quality monitoring such as inter-annotator agreement measurement, data curation, and progress monitoring. It compares the annotation disagreements only for each sentence and shows the comparison within the interface, while our system can generate a detailed disagreement report in \texttt{.pdf} file through the whole annotated content.  Besides, those web-based annotation tools need to build a server through complex configurations and some of the servers cannot be deployed on Windows systems.

The differences between \textsc{Yedda} and related work are summarised in Table \ref{tab:related}\footnote{For web-based tools, we list their server-side dependency on operating systems.}. Here ``Self Consistency'' represents whether the tool works independently or it relies on pre-installed packages. Compared to these tools, \textsc{Yedda} provides a lighter but more systematic choice with more flexibility, efficiency and less dependence on system environment for text span annotation. Besides, \textsc{Yedda} offers administrator useful toolkits for evaluating the annotation quality and analyze the detailed disagreements within annotators.

\section{\textsc{Yedda}}
\textsc{Yedda} is developed based on standard Python GUI library Tkinter\footnote{\url{https://docs.python.org/2/library/tkinter.html}}, and hence needs only Python installation as a prerequisite and is compatible with all Operating System (OS) platforms with Python installation. It offers two user-friendly interfaces for annotators and administrator, respectively, which are introduced in detail in Section \ref{sec:annotator} and Section \ref{sec:administrator}, respectively.

\subsection{Annotator Client}\label{sec:annotator}
The client is designed to accelerate the annotation process as much as possible. It supports shortcut annotation to reduce the user operation time. Command line annotation is designed to annotate multi-span in batch. In addition, the client provides system recommendations to lessen the workload of duplicated span annotation.

Figure \ref{fig:annotatorInterface} shows the interface of annotator client on an English entity annotation file. The interface consists of 5 parts. The working area in the up-left which shows the texts with different colors (blue: annotated entities, green: recommended entities and orange: selected text span). The entry at the bottom is the command line which accepts annotation command. There are several control buttons in the middle of the interface, which are used to set annotation model. The status area is below the control buttons, it shows the cursor position and the status of recommending model. The right side shows the shortcut map, where shortcut key ``a'' or ``$A$'' means annotating the text span with ``Artificial'' type and the same for other shortcut keys. The shortcut map can be configured easily\footnote{Type the corresponding labels into the entries following shortcut keys and press ``ReMap'' button.}.  Details are introduced as follows.

\subsubsection{Shortcut Key Annotation}
\textsc{Yedda} provides the function of annotating text span by selecting using mouse and press shortcut key to map the selection into a specific label. It is a common annotation process in many annotation tools \cite{stenetorp2012brat,bontcheva2013gate}. It binds each label with one custom shortcut key, this is shown in the ``Shortcuts map Labels'' part of Figure \ref{fig:annotatorInterface}. The annotator needs only two steps to annotate one text span, i.e. ``select and press''. The annotated file updates simultaneously with each key pressing process.

\subsubsection{Command Line Annotation}
\textsc{Yedda} also support the command line annotation function (see the command entry in the bottom of Figure \ref{fig:annotatorInterface}) which can execute multi-span annotation at once. The system will parse the command automatically and convert the command into multi-span annotation instructions and execute in batch. It is quite efficient for the tasks of character-based languages (such as Chinese and Japanese) with high entity density. The command follows a simple rule which is $``n1+key1+n2+key2+n3+key3+...''$, where `$n1,n2,n3$' are the length of the entities and `$key1,key2,key3$' is the corresponding shortcut key. For example, command ``$2A3D2B$'' represents annotating following 2 characters as label `$A$' (mapped into a specific label name), the following 3 characters as label `$D$' and 2 characters further as label `$B$'. 

\begin{figure}[t] 
  \centering 
  \includegraphics[width=2.5in]{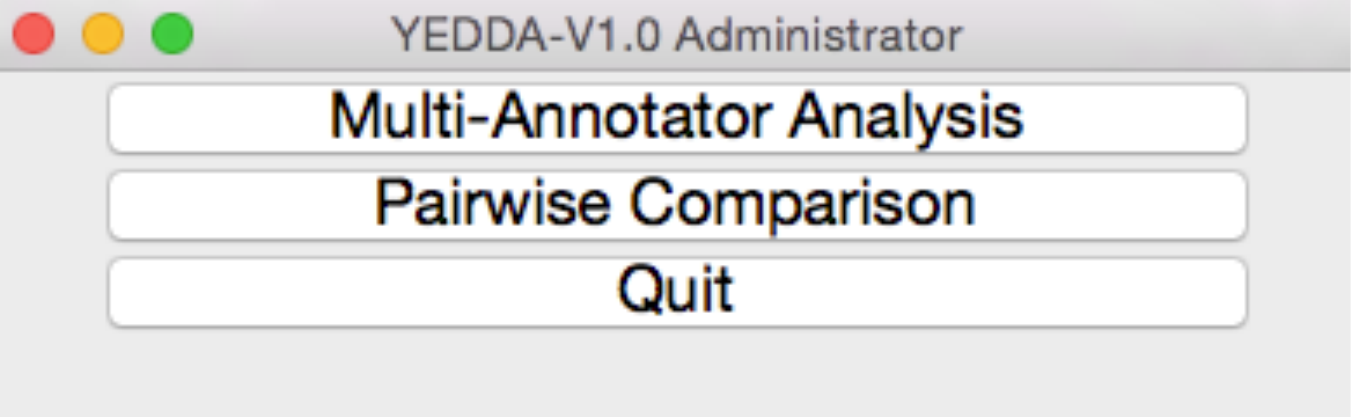}
  \caption{Administrator Interface.}
  \label{fig:adminInterface}
\end{figure}

\subsubsection{System Recommendation}
It has been shown that using pre-annotated text and manual correction increases the annotation efficiency in many annotation tasks \cite{meurs2011towards,stenetorp2012brat}. \textsc{Yedda} offers annotators with system recommendation based on the existing annotation history. The current recommendation system incrementally collects annotated text spans from sentences that have been labeled, thus gaining a dynamically growing lexicon. Using the lexicon, the system automatically annotates sentences that are currently being annotated by leveraging the forward maximum matching algorithm. The automatically suggested text spans and their types are returned with colors in the user interface, as shown in green in Figure \ref{fig:annotatorInterface}. Annotators can use the shortcut to confirm, correct or veto the suggestions. The recommending system keeps online updating during the whole annotation process, which learns the up-to-date and in-domain annotation information. The recommending system is designed as ``pluggable'' which ensures that the recommending algorithm can be easily extended into other sequence labeling models, such as Conditional Random Field (CRF)\footnote{Those sequence labeling models work well on big training data. For limited training data, the maximum matching algorithm gives better performance.} \cite{lafferty2001conditional}. The recommendation can be controlled through two buttons ``RMOn'' and ``RMOff'', which enables and disables the recommending function, respectively.  

\begin{figure}[t] 
  \centering 
  \includegraphics[width=\columnwidth]{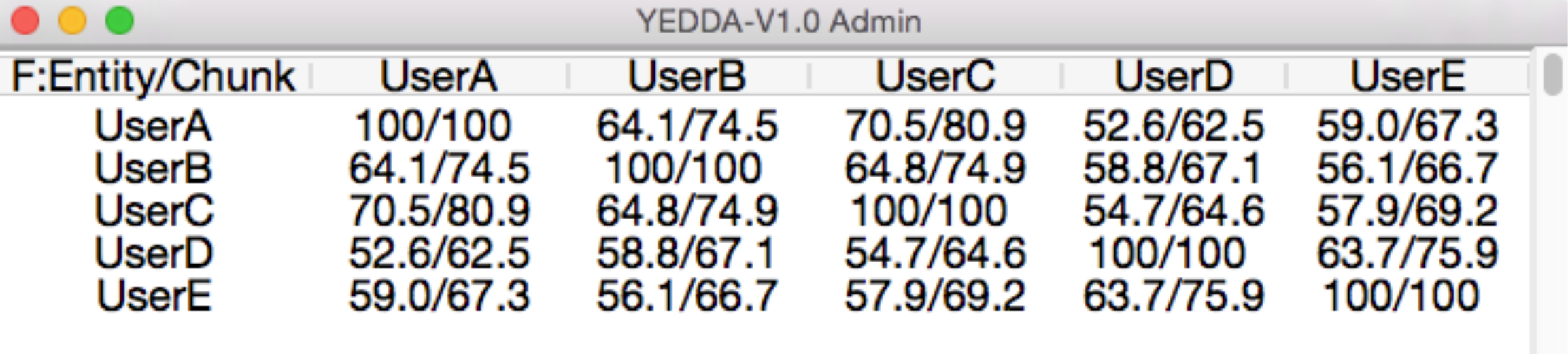}
  \caption{Multiple annotators F1-score matrix.}
  \label{fig:matrix}
\end{figure}

\subsubsection{Annotation Modification}
It is inevitable that the annotator or the recommending system gives incorrect annotations or suggestions. Based on our annotation experience, we found that the time cost of annotation correction cannot be neglected. Therefore, \textsc{Yedda} provides several efficient modification actions to revise the annotation:

\noindent \textbullet $\;$ \textbf{Action withdraw}: annotators can cancel their previous action and let system return to the last status by press the shortcut key \texttt{Ctrl+z}.

\noindent \textbullet $\;$ \textbf{Span label modification}: if the selected span has the correct boundary but receives an incorrect label, annotator only needs to put the cursor inside the span (or select the span) and press the shortcut key of the right label to correct label. 

\noindent \textbullet $\;$ \textbf{Label deletion}: similar to the label modification, the annotator can put the cursor inside the span and press shortcut key \texttt{q} to remove the annotated (recommended) label.

\subsubsection{Export Annotated Text}
As the annotated file is saved in \texttt{.ann} format, \textsc{Yedda} provides the ``Export'' function which exports the annotated text as standard format (ended with \texttt{.anns}). Each line includes one word/character and its label, sentences are separated by an empty line. The exported label can be chosen in either BIO or BIOES format \cite{ratinov2009design}.

\subsection{Administrator Toolkits}\label{sec:administrator}
For the administrator, it is important and necessary to evaluate the quality of annotated files and analyze the detailed disagreements of different annotators. Shown in Figure \ref{fig:adminInterface}, \textsc{Yedda} provides a simple interface with several toolkits for administrator monitoring the annotation process.

\subsubsection{Multi-Annotator Analysis}
To evaluate and monitor the annotation quality of different annotators, our Multi-Annotator Analysis (MAA) toolkit imports all the annotated files and gives the analysis results in a matrix. As shown in Figure \ref{fig:matrix}, the matrix gives the F1-scores in full level (consider both boundary and label accuracy) and boundary level (ignore the label correctness, only care about the boundary accuracy) of all annotator pairs. 

\begin{figure}[t] 
  \centering 
  \fbox{\includegraphics[width=3in]{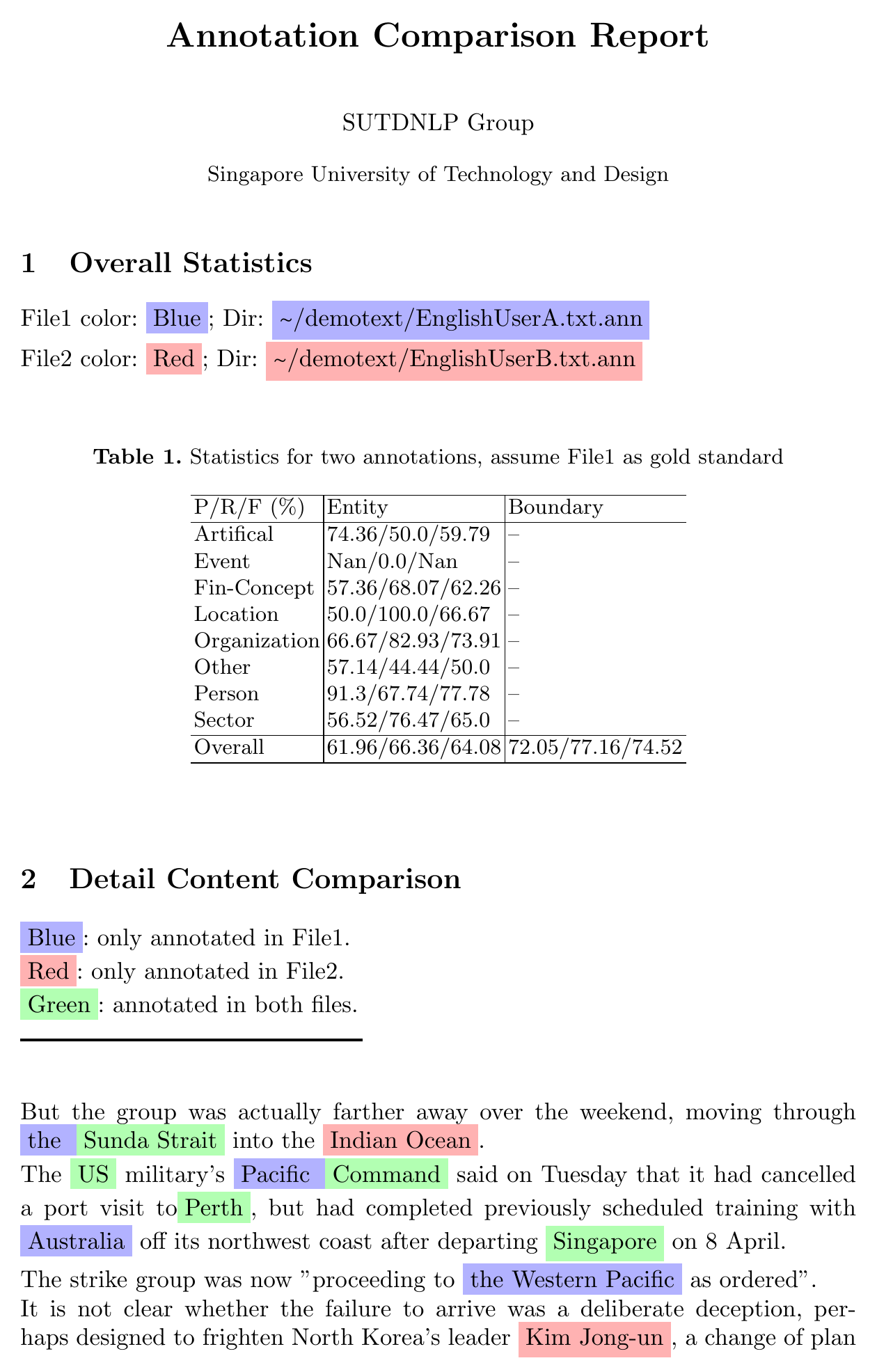}}
  \caption{Detailed report for annotator pair.}
  \label{fig:detail}
\end{figure}

\subsubsection{Pairwise Annotators Comparison}
If an administrator wants to look into the detailed disagreement of annotators, it is quite convenient by using the Pairwise Annotators Comparison (PAC). PAC loads two annotated files and generates a specific comparison report file\footnote{The report is generated in \texttt{.tex} format and can be complied into \texttt{.pdf} file.} for the two annotators. As shown in Figure \ref{fig:detail}, the report is mainly in two parts:

\noindent \textbullet $\;$ \textbf{Overall statistics}: it shows the specific precision, recall and F1-score\footnote{Notice that we assume ``File1'' as a gold standard. This only affects the order of precision and recall, while the F1-score keeps same if we choose the other file as gold standard.} of two files in all labels. It also gives the three accuracy indexes on overall full level and boundary level in the end.

\noindent \textbullet $\,$ \textbf{Content comparison}: this function gives the detailed comparison of two annotated files in whole content. It highlights the annotated parts of two annotators and assigns different colors for the agreed and disagreed span.

\begin{figure}[t] 
  \centering 
  \includegraphics[width=3in]{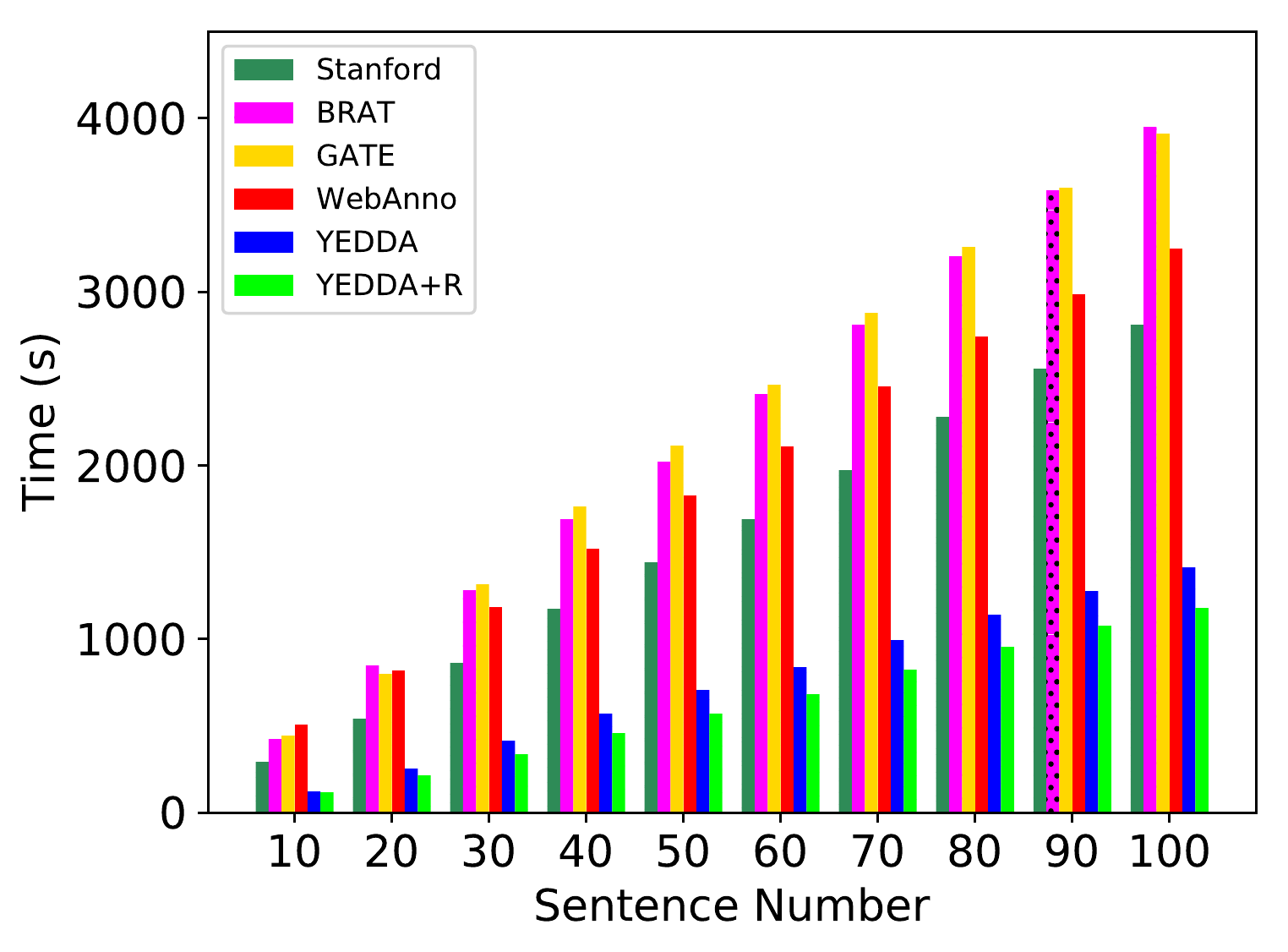}
  \caption{Speed comparison.}
  \label{fig:speed}
\end{figure}

\section{Experiments}
Here we compare the efficiency of our system with four widely used annotation tools. We extract 100 sentences from CoNLL 2003 English NER \cite{tjong2003introduction} training data, with each sentence containing at least 4 entities. Two undergraduate students without any experience on those tools are invited to annotate those sentences\footnote{We ask the students to annotate those sentences several rounds to get familiar with the entities before they start the final exercise with recording.}. Their average annotation time is shown in Figure \ref{fig:speed}, where ``\textsc{Yedda}+R'' suggests annotation using \textsc{Yedda} with the help of system recommendation. The inter-annotator agreements for those tools are closed, which around 96.1\% F1-score. As we can see from the figure, our \textsc{Yedda} system can greatly reduce the annotation time. With the help of system recommendation, the annotation time can be further reduced. We notice that ``\textsc{Yedda}+R'' has larger advantage with the increasing numbers of annotated sentences, this is because the system recommendation gives better suggestions when it learns larger annotated sentences. The ``\textsc{Yedda}+R'' gives 16.47\% time reduction in annotating 100 sentences\footnote{The speed improvement by recommendation depends on the density of text spans. We suggest enabling the recommendation model in the task whose text contains dense and recurring text spans.}.

\section{Conclusion and Future Work}
We have presented a lightweight but systematic annotation tool, \textsc{Yedda}, for annotating the entities in text and analyzing the annotation results efficiently. In order to reduce the workload of annotators, we are going to integrate active learning strategy in our system recommendation part in the future. A supervised sequence labeling model (such as CRF) is trained based on the annotated text, then unannotated sentences with less confidence (predicted by this model) are reordered in the front to ensure annotators only annotate the most confusing sentences.

\section{Acknowledgements}

We thank Yanxia Qin, Hongmin Wang, Shaolei Wang, Jiangming Liu, Yuze Gao, Ye Yuan, Lu Cao, Yumin Zhou and other members of SUTDNLP group for their trials and feedbacks.  Yue Zhang is the corresponding author. Jie is supported by the YEDDA grant 52YD1314.

\bibliography{acl2018}
\bibliographystyle{acl_natbib}

\end{document}